\pdfoutput=1

\documentclass[11pt]{article}
\usepackage{todonotes}
\usepackage{acl}
\usepackage{tabularx}
\usepackage{times}
\usepackage{latexsym}

\usepackage[T1]{fontenc}

\usepackage[utf8]{inputenc}

\usepackage{microtype}

\usepackage{float}

\title{Searching for PETs: Using Distributional and Sentiment-Based Methods to Find Potentially Euphemistic Terms }

\author{Patrick Lee \and Martha Gavidia \and Anna Feldman \and Jing Peng \\
         Montclair State University \\ Montclair, New Jersey \\ \{leep6, gavidiam1 feldmana, pengj\}@montclair.edu\\}

\begin{document}
\maketitle
\begin{abstract}
This paper presents a linguistically driven proof of concept for finding \emph{potentially euphemistic terms}, or PETs. Acknowledging that PETs tend to be commonly used expressions for a certain range of sensitive topics, we make use of distributional similarities to select and filter phrase candidates from a sentence and rank them using a set of simple sentiment-based metrics. We present the results of our approach tested on a corpus of sentences containing euphemisms, demonstrating its efficacy for detecting single and multi-word PETs from a broad range of topics. We also discuss future potential for sentiment-based methods on this task. 
\end{abstract}

\section{Introduction}
Euphemisms are mild or indirect expressions used in place of harsher or more offensive ones. They can be used to show politeness when discussing sensitive or taboo topics  \cite{bakhriddionova2021needs} such as saying \emph{passed away} instead of \emph{died}, or as a way to make unpleasant or unappealing things sound better \cite{karam2011truths}, such as \emph{ethnic cleansing} instead of \emph{genocide}.  They can even be used as a means to conceal the truth \cite{rababah2014translatability}; for example, saying \emph{enhanced interrogation techniques} but meaning \emph{torture}. Euphemisms pose a challenge to natural language processing due to this figurative behavior, but also because they can have a literal interpretation in certain contexts. Furthermore, humans may not agree on what a euphemism is \cite{gavidia-etal:2022}. Thus, we consider any words/phrases used in this nature to be a \emph{Potentially Euphemistic Term} (PET).



In this paper, we present a proof of concept for finding PETs in an input sentence, and apply it to a novel euphemism corpus \cite{gavidia-etal:2022}\footnote{Our code is available at \url{https://github.com/marsgav/PETDetection}.}. We base our approach on several linguistic intuitions: (1) PETs tend to be commonly used expressions about a certain range of sensitive topics, (2) humans make a conscious lexical choice to convey politeness and formality and (3) because of their linguistic function, PETs should result in greater sentiment shifts when replaced by their literal interpretations; we experiment with distributionally similar alternatives as a source of such interpretations. Leveraging a variety of existing tools (Gensim's Phrases \cite{rehurek2011gensim}, word2vec classes \cite{mikolov2013efficient}, and roBERTa \cite{liu2019roberta}), we implement a simple algorithm to extract, filter, and rank PET candidates. Despite its simplicity, our approach is able to identify the target euphemism as one of the top two phrase candidates for 725 out of 1382 sentences in our test dataset. It also shows promising results in identifying PETs that were not originally marked, as well as for sentences outside our dataset. We believe our results and subsequent discussion are an important baseline for using distributional and sentiment-based methods for detecting euphemisms.

The structure of the paper is as follows: in Section 2, we discuss related work surrounding euphemisms. Section 3 provides details on the text data used, Section 4 describes our approach broken down into 4 stages: \textit{phrase extraction, phrase filtering, phrase paraphrasing and phrase ranking}. Section 5 includes our results and a quantitative and qualitative analysis, and Section 6 concludes with future work.

\section{Related Work}

Computational approaches to processing euphemisms \cite{felt2020recognizing,zhu2021self,zhu-bhat-2021-euphemistic-phrase,magu2018determining, kapron-king-xu-2021-diachronic,gavidia-etal:2022} have shown much promise, but the dynamic nature of euphemisms remains an obstacle.  A euphemism annotation task conducted by \citet{gavidia-etal:2022} shows that the inherent ambiguity of euphemisms leads to low agreement in what qualifies as a euphemism.  Through this task, the researchers found that some euphemisms are used so often to discuss sensitive topics (e.g., venereal disease as a euphemism for sexually transmitted disease), that they become \emph{commonly accepted terms}, or CATs.  Additionally, they find that even when annotators agreed on the intended meaning of a euphemism, e.g. \textit{slim} as a euphemism for \textit{skinny}, they still did not agree on the label of euphemistic vs. non euphemistic.  The nuance associated with euphemisms still remains one of the biggest challenges.

\citet{felt2020recognizing} were one of the first to tackle euphemisms from a computational standpoint. They leverage sentiment analysis to recognize \emph{x-phemisms}, which is the term they use to refer to both euphemisms and dysphemisms.  Whereas euphemisms are polite expressions to discuss sensitive topics, dysphemisms are purposely direct, blunt and can be derogatory.  They find near-synonym pairs for three topics: lying, firing and stealing, and use a weakly supervised bootstrapping algorithm for semantic lexicon induction \cite{thelen2002bootstrapping}.  They use lexical cues and sentiment analysis to classify phrases as euphemistic, dysphemistic or neutral. Their approach is interesting, as it is the first of its kind and their use of sentiment analysis to identify euphemisms has inspired our work.

\citet{zhu2021self} approach the task of discovering euphemisms from the lens of content moderation. Their goal was the detection of euphemisms used for formal drug names on social media. They define two problems: the first is the detection of euphemisms, and the second is identifying what the euphemisms found actually refer to. However, their view on euphemisms is different from ours, as they treat euphemisms simply as code words. This work is similar to \citet{magu2018determining}, who also explore euphemisms as code words in hate speech.  \citet{zhu-bhat-2021-euphemistic-phrase} and \citet{zhu2021self} both treat detection and identification as a masked language problem where they use a masked language model (MLM) as a filter to get rid of sentences that are not related to their seedlist of euphemisms and then again to find euphemistic candidates.  Like \citet{felt2020recognizing}, \citet{zhu2021self} and \citet{zhu-bhat-2021-euphemistic-phrase} show promise, though their narrow topic focus limit the kinds of euphemisms that can be found.

Lastly, \citet{kapron-king-xu-2021-diachronic} conduct a diachronic evaluation of euphemism usage between genders. While this work is not aimed at finding euphemisms, their work provides many of the PETs used in the creation of the Euphemism Corpus \cite{gavidia-etal:2022}, which we use in this paper.

\section{Data}
Our work utilizes a Euphemism Corpus created by \cite{gavidia-etal:2022} as our test data.  The raw text data for this corpus comes from The Corpus of Global Web-Based English (GloWbE)\cite{davies2015expanding}. GloWbE contains text data for 20 English speaking countries from websites, blogs and forums; this corpus is compiled using just a portion of the US Dialect of English text. \\
The Euphemism Corpus contains 1,382 euphemistic sentences, each annotated with one potentially euphemistic term per sentence.  These potentially euphemistic terms, or PETs \cite{gavidia-etal:2022} are single and multi word expressions that are used in a euphemistic sense.\\
Futhermore, we use the US Dialect of English portion of GloWbE to train a Phrases model (gensim) \cite{rehurek2011gensim} to create word collocations within our data which are then fed into a word2vec model to produce vector representations for the words in our corpus.  The following section explains both of these aspects in further detail. 

\section{Our Approach}
The algorithm developed for this experiment performs the following sub tasks to identify a PET in a sentence:  \textit{phrase extraction, phrase filtering, phrase paraphrasing and phrase ranking}.  Simply put, the algorithm locates all of the single and multi word expressions within a sentence and through the subsequent tasks, determines which expressions may be a PET.  

\subsection{Phrase Extraction}
We use the phrase (collocation) detection model, Phrases, in the Gensim library \cite{rehurek2011gensim} to identify single and multi word expressions within the US Dialect of English portion of GloWbE \cite{davies2015expanding}.  Phrases takes raw text as input and detects a bigram if a scoring function for two words exceeds a certain threshold.  It joins two unigrams into a single token, separated by an underscore. We use Phrases to train our data twice in order to create up to 3 word expressions to account for PETs like \textit{enhanced interrogation techniques}. Upon training, Phrases creates a Phraser object that can be applied to new text data to identify bigram and trigram expressions.  As such, we use this Phraser object on the Euphemism Corpus, resulting in identification of single and multiword expressions contained within it.

\subsection{Phrase Filtering}
The single and multiword expressions found with Phrases now need to be topically filtered. This step is essential in identifying the phrases that are related to a sensitive topic. We remove all stopwords, and then, leveraging the embeddings created with word2vec, calculate the cosine similarity between the phrases and a list of words representing sensitive topics \cite{gavidia-etal:2022}. These sensitive topics include: death, sexual activity, employment, bodily functions, politics, physical/mental attributes, and substances. We notice that many of the PETs in the Euphemism Corpus have a summed cosine similarity score above 1.5; therefore, we empirically set this as the threshold. Every phrase with a similarity measure above this is referred to as a \textit{quality phrase} and moves on to the next task of paraphrasing.

\subsection{Phrase Paraphrasing}
The idea behind paraphrasing a PET is that, in theory, if we replace quality phrases with "paraphrases" that are more literal, there should be a shift in the sentiment of the sentence.  Since using euphemisms can be seen as a conscious lexical choice made to avoid awkward or uncomfortable situations, when we choose to use a PET, our goal is to make our speech less negative, more positive and less offensive.  
We test this by "paraphrasing" quality phrases using the top 25 most similar words as output by word2vec (excluding paraphrases which contain the quality phrase as a substring, as these are not really distinct alternatives) and perform sentiment analysis to measure negative, positive and offensive scores\cite{liu2019roberta} before and after replacement.\\
Using the distributionally similar words output by word2vec follows the intuition that phrase semantics are determined by their context, and that phrases which have the same mentions should have the same semantics \cite{li2022uctopic}.  We recognize that these are not official paraphrases; however, as seen by the example below for the PET \emph{intoxicated}, word2vec produces good results.\\

\includegraphics[width=0.98\columnwidth]{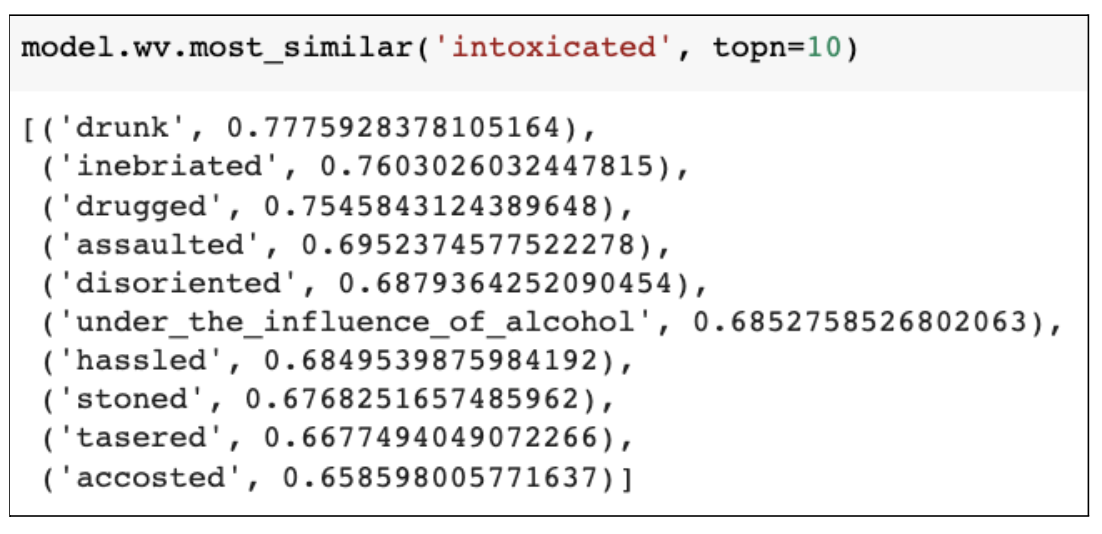}

From this list, we see that "drunk" and "under the influence of alcohol" would be considered literal interpretations of "intoxicated", and as such, would serve as suitable replacements for the paraphrasing task.
\subsection{Phrase Ranking}
To measure differences in sentiment and offensiveness of the original sentences before and after substituting with alternatives output by word2vec, we use a roBERTa base model trained on tweets for sentiment analysis and offensive language identification \cite{liu2019roberta}. We chose RoBERTa's sentiment and offensiveness models because they have been shown to be useful in distinguishing PETs from other phrases \cite{gavidia-etal:2022}. The specific scores we utilize are negative, neutral, and positive sentiment scores, as well as non-offensiveness and offensiveness scores. We calculate scores for all replacements and aggregate them into a single score as a measure of which PET had replacements that caused the greatest shift in sentiment. Reasoning that alternatives to PETs are likely more polarized than alternatives to non-PETs, we rank the quality phrases using this aggregate from highest to lowest. The phrases with the top 2 highest scores in each sentence are deemed to be PET candidates. 

Empirically, we notice that both offensiveness scores tend to be particularly useful for distinguishing euphemisms from polarized (but otherwise non-euphemistic) terms, so we attribute more weight to them. We hypothesize that both non-offensive and offensive scores are useful because terms that are distributionally similar to PETs are likely to be either (1) similar, non-offensive alternatives or (2) their offensive alternatives. See Appendix~\ref{sec:sentshifts} for an illustration of the paraphrasing stage, along with sample sentiment shifts.

\section{Results and Discussion}
This section provides our quantitative and qualitative analyses and a discussion on the failures and limitations of our algorithm.

\subsection{Quantitative Analysis}
Table \ref{subtask-performance} summarizes the results from each step of our procedure.  The second column shows the number of total candidate phrases at every stage while the last column shows how many test sentences, out of 1382, still retain the target PETs in the list of candidates at that stage. Note the paraphrasing stage shows no changes as this stage is not meant to reduce the list of PETs. 

\begin{table}[H]
\begin{center}
\begin{tabular}{|p{0.45\columnwidth}|p{0.2\columnwidth}|p{0.2\columnwidth}|}
\hline
\textbf{Stage} & \textbf{\# Candidates} & \textbf{\# Targets Retained}\\
\hline
Phrase Extraction & 31348 & 1251 \\
Phrase Filtering & 10503 & 1198 \\
Phrase Paraphrasing & 10503 & 1198 \\
Phrase Ranking & 2728 & \textbf{725} \\
\hline
\end{tabular}
\caption{\label{subtask-performance}
A summary of the subtasks in our algorithm, along with the number of candidate phrases and PETs that were retained after each.
}
\end{center}
\end{table}
The algorithm correctly identifies the target PET in 725 sentences.  Additionally, through human evaluation, we find that it also identifies new non-target PETs in the data. Out of the 725 PETs deemed to have been successfully detected, 468 of them were ranked as the 1st place candidate, while 257 were 2nd place. Overall, this gives us a success rate of about \textbf{52.5\%}. Since there was an average of \textbf{7.6} phrase candidates per sentence, we calculate the chance of randomly selecting the target to be one of the top two candidates to be 2 * (1/7.6) $\approx \textbf{26.3\%}$. The sizable improvement over this baseline — which doesn't include new, non-target PETs that were detected — leads us to believe our results are significant.  

\subsection{Qualitative Analysis}
Below, Table \ref{subtask-performance} includes an example of a correctly identified target PET as well as a new PET that was not annotated for in our test data.  While the target PET \textit{mentally disabled} is identified as the second top ranked phrase, we deem the first ranked phrase, \textit{intoxicated person}, to be a PET as well.  

\begin{table}[h!]
\begin{tabularx}{\columnwidth}{X}
\textbf{Sentence:} \textit{in addition bats that are found in a room with a person who can not reliably rule out physical contact for example a sleeping person a child a mentally disabled person or an intoxicated person will need to be tested for rabies} \\
\textbf{Target PET:} mentally disabled \\
\textbf{ExtractedPhrases:} {[}'in', 'addition', 'bats', 'that', 'are', 'found', 'in', 'a', 'room', 'with', 'a', 'person\_who', 'can\_not', 'reliably', 'rule\_out', 'physical\_contact', 'for', 'example', 'a', 'sleeping', 'person', 'a', 'child', 'a', 'mentally\_disabled', 'person', 'or', 'an', 'intoxicated\_person', 'will\_need', 'to', 'be\_tested', 'for', 'rabies'{]} \\
\textbf{QualityPhrases:} {[}'bats', 'person\_who', 'can\_not', 'reliably', 'physical\_contact', 'sleeping', 'person', 'child', 'mentally\_disabled', 'intoxicated\_person', 'be\_tested', 'rabies'{]}\\
\textbf{RankedPhrases:} {[}(\underline{'intoxicated person'}, 2.898948520421982), (\underline{'mentally disabled'}, 2.7745959013700485), ('rabies', 2.036529041826725), ('physical contact', 1.7015496864914894), ('can not', 1.6931570619344711), ('sleeping', 1.267698973417282), ('person', 1.171182319521904), ('person who', 1.0447067320346832), ('bats', 0.9130769670009613), ('be tested', 0.864994041621685), ('reliably', 0.8625116124749184), ('child', 0.23687118291854858){]}\\

\end{tabularx}
\caption{Example of target PET 'mentally disabled' as second ranked phrase with new PET 'intoxicated person' ranked first. }
\end{table}

We include additional examples of sentences in which the target PET was correctly identified as a top two candidate phrase in Appendix~\ref{sec:successpets}. Appendix~\ref{sec:newpets} also showcases more new PETs that were found - by human evaluation. We discuss instances where our algorithm failed to detect a target PET in the following section.  

\subsection{Failures}
The output candidates may not include the target PET for a couple of reasons: (1) it is not retained from the phrase detection or topic filtering stages, or (2) it produces a low sentiment or offensiveness shift compared to other candidates. Notably, for (1), we notice MWEs are sometimes not collocated properly, either because they aren't detected as a common collocation (e.g., 'between' and 'jobs' are never joined into a single phrase) or because they are collocated with other terms (e.g., 'almost\_lost' and 'my\_lunch' are detected to be MWEs, but as a result, not 'lost\_my\_lunch'). For (2), we notice that other candidates (polarized phrases or broad nouns in particular) simply produce higher shifts in all or most sentiment categories compared to the target PET. (See Appendix~\ref{sec:failures} for more examples.) As such, while simply computing the increases in sentiment scores and prioritizing offensiveness scores produces workable results for this proof of concept, there is a clear need to experiment with better methods for utilizing sentiment; this is left to future experimentation.

\section{Conclusion and Future Work}

Our work is a proof of concept for finding PETs in a given euphemistic sentence. While our algorithm produces significant results, we recognize the limitations of our work and propose the following ideas for advancement of this specific task. Firstly, we rely on the Gensim library for identifying multi-word expressions and obtaining word embeddings, but experimentation with different parameters and techniques (e.g., using different phrase extraction methods, different bigram scoring functions or contextualized word embeddings) may yield better results. Secondly, a mechanism for filtering each candidate's alternatives could help reduce the number of semantically dissimilar replacements during the paraphrasing stage. Next, while we only use aggregate increases in sentiment and offensiveness scores for ranking candidates, a variety of other methods (e.g., taking averages or maximums) and measures (e.g., indirectness and vagueness) may be useful for distinguishing PETs. Lastly, while the task of differentiating literal versus euphemistic usages of PETs is not a focus on this paper, our algorithm shows some promise on the issue (see Appendix~\ref{sec:newapps}), and it is an important task that could use future work; Appendix~\ref{sec:newapps} also shows the performance of our algorithm on unseen data.

\section*{Acknowledgements}
This material is based upon work supported by the National Science Foundation under Grant No.~1704113.
\bibliography{anthology,custom}
\bibliographystyle{acl_natbib}

\newpage
\onecolumn
\appendix

\newpage
\section{Example Sentiment Shifts when Replacing PET Candidates}
\label{sec:sentshifts}

Below we illustrate the paraphrasing stage for two sample PETs (only showing the top 10 replacements for each). Each replacement is listed along with the sentiment shifts it produces in the original sentence (of which, only the increases are aggregated into a final score for the candidate). The five numbers indicate, in order, the [negative, neutral, positive, non-offensive, offensive] sentiment shifts.

\begin{table}[h!]
\begin{tabular}{|p\textwidth|}
\hline
\textbf{\textbf{Example 1}} \\ \hline
\textbf{Original Sentence:} the city has told quite a few mistruths in order to get this city office approved \\
\textbf{Target PET:} mistruths \\
\textbf{Top 10 Replacements:}  \\
half-truths
[-0.04721798, 0.02886498, 0.018352773, -0.01894176, 0.018941715] \\
outright lies
[0.5365411, -0.49552178, -0.041019425, -0.19229656, 0.1922966] \\
falsehoods
[0.4495571, -0.4112787, -0.0382785, -0.117421865, 0.11742185] \\
untruths
[0.15190402, -0.13002646, -0.021877721, -0.03839171, 0.03839159] \\
half truths
[0.12055096, -0.11013514, -0.010415968, -0.029488266, 0.029488236] \\
blatant lies
[0.58376455, -0.54074687, -0.04301778, -0.268206, 0.268206] \\
race-baiting
[0.08099219, -0.06833702, -0.01265521, -0.05247575, 0.052475646] \\
spewing lies
[0.5345895, -0.49333698, -0.041252725, -0.34204996, 0.34205] \\
lies and distortions
[0.5648471, -0.5210455, -0.043801878, -0.15087801, 0.15087792] \\
fearmongering
[0.12043443, -0.107453406, -0.012981113, -0.029256463, 0.029256403] \\
\textbf{Comment:} Note other potentially non-offensive alternatives like "half-truths" and "untruths" (which sometimes result in greater shifts in non-offensiveness than this example), and literal interpretations like "outright lies" and "lies and distortions" (which result in significant offensiveness shifts).  \\ \hline
\end{tabular}
\end{table}

\begin{table}[h!]
\begin{tabular}{|p\textwidth|}
\hline
\textbf{\textbf{Example 2}} \\ \hline
\textbf{Original Sentence:} after deadly ethnic riots rocked southern kyrgyzstan last month one georgian minister claimed that russia has been behind the ethnic cleansing of uzbeks \\
\textbf{Target PET:} ethnic cleansing \\
\textbf{Top 10 Replacements:}  \\
genocide
[0.022250175, -0.021986336, -0.00026384578, -0.017507195, 0.017507195]\\
collective punishment
[-0.027520716, 0.026981518, 0.0005392225, 0.0048098564, -0.004809916]\\
apartheid
[-0.00591588, 0.005712375, 0.000203504, 0.011624634, -0.011624634]\\
massacres
[0.0055012107, -0.0054178983, -8.32919e-05, -0.004523158, 0.004523158]\\
israeli occupation
[-0.019839048, 0.019360632, 0.00047835405, 0.039074123, -0.039074093]\\
islamic terrorism
[-0.004287958, 0.0042657405, 2.238946e-05, -0.028740644, 0.028740555]\\
mass murder
[0.021661818, -0.021403424, -0.00025822758, -0.08434576, 0.08434579]\\
sectarian conflict
[-0.02048409, 0.020332336, 0.00015191245, 0.047309637, -0.047309637]\\
islamization
[-0.035422206, 0.03482026, 0.00060210144, 0.027191758, -0.027191669]\\
foreign occupation
[-0.018171906, 0.017809838, 0.0003620449, 0.04308176, -0.0430817]\\
\textbf{Comment:} Note the literal interpretations "genocide", "massacres" and "mass murder".  \\ \hline
\end{tabular}
\end{table}

\newpage
\section{Examples of Successfully Detected PETs}
\label{sec:successpets}
Below are examples where our algorithm successfully detected the target PET.  The output is as follows. We identify a Target PET along with the sentence it belongs to.  The first set of phrases, \textit{ExtractedPhrases}, are those retrieved through Phrases; after using word2vec to further filter phrases according to our topics, we obtain our \textit{QualityPhrases}; finally, we display our \textit{RankedPhrases} where our candidate PETs appeared in one of the top two rankings.
\begin{table}[h!]
\begin{tabular}{|p\textwidth|}
\hline
\textbf{\textbf{Example 1}} \\ \hline
\textbf{Target PET:} psychiatric hospital \\
\textbf{Sentence:} \textit{you may believe that if you have signed yourself voluntarily into a psychiatric hospital you can sign yourself out and leave when you decide to do so} \\ 
\textbf{ExtractedPhrases:} {[}'you\_may', 'believe\_that', 'if\_you', 'have', 'signed', 'yourself', 'voluntarily', 'into', 'a', 'psychiatric\_hospital', 'you\_can', 'sign', 'yourself', 'out', 'and', 'leave', 'when\_you', 'decide', 'to', 'do', 'so'{]} \\ 
\textbf{QualityPhrases:} {[}'believe\_that', 'if\_you', 'voluntarily', 'psychiatric\_hospital', 'sign', 'leave', 'when\_you', 'decide'{]} \\ 
\textbf{RankedPhrases:} {[}(\underline{'psychiatric hospital'}, 7.978855848312378), ('voluntarily', 4.409763276576996), ('sign', 2.7386385649442673), ('if you', 2.3423103243112564), ('believe that', 1.915013164281845), ('when you', 1.7038534581661224), ('leave', 1.6538356691598892), ('decide', 1.548440158367157){]} \\ \hline
\end{tabular}
\end{table}
\begin{table}[h!]
\begin{tabular}{|p\textwidth|}\hline
\textbf{Example 2} \\ \hline
\textbf{Target PET:} armed conflict \\ 
\textbf{Sentence:} \textit{when this happens something of considerable legal significance does occur the law of armed conflict begins to govern belligerent relations between the states} \\ 
\textbf{ExtractedPhrases: }{[}'when', 'this\_happens', 'something', 'of', 'considerable', 'legal\_significance', 'does\_occur', 'the', 'law', 'of', 'armed\_conflict', 'begins', 'to', 'govern', 'belligerent', 'relations\_between', 'the', 'states'{]} \\ 
\textbf{QualityPhrases:} {[}'this\_happens', 'considerable', 'legal\_significance', 'does\_occur', 'law', 'armed\_conflict', 'govern', 'belligerent', 'relations\_between'{]} \\
\textbf{RankedPhrases:} {[}('legal significance', 3.8234215676784515), (\underline{'armed conflict'}, 3.674671307206154), ('this happens', 3.6536989957094193), ('belligerent', 2.823164239525795), ('considerable', 1.5059781521558762), ('govern', 1.2904964834451675), ('does occur', 1.1230540722608566), ('relations between', 0.7008794546127319), ('law', 0.5298605561256409){]} \\
 \hline
\end{tabular}
\end{table}
\begin{table}[h!]
\begin{tabular}{|p\textwidth|}\hline
\textbf{Example 3} \\ \hline
\textbf{Target PET:} pro-life \\ 
\textbf{Sentence:} \textit{however i am also a person who respects life in all of its forms and so i could also qualify as a pro-life person} \\ 
\textbf{ExtractedPhrases:} {[}'however\_i\_am', 'also', 'a', 'person\_who', 'respects', 'life', 'in', 'all', 'of', 'its\_forms', 'and', 'so', 'i\_could', 'also', 'qualify\_as', 'a', 'pro-life', 'person'{]} \\ 
\textbf{QualityPhrases:} {[}'person\_who', 'life', 'its\_forms', 'qualify\_as', 'pro-life', 'person'{]} \\
\textbf{RankedPhrases:} {[}(\underline{'pro-life'}, 14.923447516746819), ('person', 4.519588744267821), ('qualify as', 2.345528486184776), ('life', 1.7386144306510687), ('its forms', 1.536714962683618), ('person who', 1.4028910771012306){]} \\ \hline
\end{tabular}
\end{table}

\newpage
\section{Examples of New PETs Found}
\label{sec:newpets}
Since our algorithm works by placing a candidate PET in one of the top two rankings, we evaluated the results and found that new PETs were found and correctly placed in top ranking positions.  One of the limitations of the Euphemism Corpus is that it only includes one annotated PET per sentence, our algorithm shows potential to expand upon the annotations in the corpus to include the new PETs found.  We underline the new PETs in the examples below as well as provide our interpretations.

\begin{table}[h!]
\begin{tabular}{|p{\textwidth}|}\hline
\textbf{Example 1} \\\hline
\textbf{Sentence:} \textit{or acknowledge real-world trade-offs such as the strong likelihood of amount of of civilian casualties if aq detainees were treated according to either geneva convention or uc criminal law standards} \\
\textbf{RankedPhrases:} {[}(\underline{'civilian casualties'}, 3.9511645138263702), ('criminal law', 2.082954853773117), ('trade-offs', 2.0316174626350403), ('geneva convention', 1.7544293403625488), ('acknowledge', 1.5634678304195404), ('detainees were', 1.2355359494686127), ('treated', 1.2001541256904602), ('standards', 0.8081734478473663){]} \\
\textbf{New PET:} civilian casualties\\
\textbf{Interpretation:} the unintended deaths of civilians \\ \hline
\end{tabular}
\end{table}

\begin{table}[h!]
\begin{tabular}{|p{\textwidth}|}
\hline
\textbf{Example 2 }\\ \hline
\textbf{Sentence:} \textit{pelosi says she was briefed by bush administration officials on the legal justification for using waterboarding but that they never followed through on promises to inform her when they actually began using enhanced interrogation techniques} \\
\textbf{RankedPhrases:} {[}(\underline{'using waterboarding'}, 6.236076384782791), ('enhanced interrogation techniques', 3.640248477458954), ('she was', 1.1687388718128204), ('legal justification', 1.1285315454006195), ('when they', 0.9696991741657257){]} \\
\textbf{New PET:} using waterboarding \\
\textbf{Interpretation:} a form of torture where a person is strapped down to a board and water is poured over their face in a way that is similar to drowning \\ \hline
\end{tabular}
\end{table}

\begin{table}[h!]
\begin{tabular}{|p{\textwidth}|}
\hline
\textbf{Example 3} \\ \hline
\textbf{Sentence:} \textit{religious people often complain that secular therapists see their faith as a problem or a symptom rather than as a conviction to be respected and incorporated into the therapeutic dialogue a concern that is especially pronounced among the elderly and twentysomethings} \\
\textbf{RankedPhrases:} {[}(\underline{'secular therapists'}, 1.9648141264915466), ('especially pronounced', 1.7061323672533035), ('among the elderly', 1.6529535502195358), ('their faith', 1.3943422138690948), ('rather than', 1.2891167849302292), ('concern', 1.2376690953969955), ('religious people', 0.9915256798267365), ('symptom', 0.8965674340724945), ('therapeutic', 0.8766119182109833), ('twentysomethings', 0.8552953451871872), ('be respected', 0.5095183551311493), ('conviction', 0.48858143389225006), ('dialogue', 0.3565850257873535){]} \\
\textbf{New PET:} secular therapists \\
\textbf{Interpretation:} a non-religious therapist who uses science based therapy methods \\ \hline
\end{tabular}
\end{table}

\newpage
\section{Examples of Failed Target PET Detection}
\label{sec:failures}
The following examples show instances where our algorithm failed to correctly detect the target PET.  We include examples showing sentences in which our MWE extraction method failed to initially recognize a PET as a phrase, and other examples showing where different words, such as action words, had a higher ranking.

\begin{table}[h!]
\begin{tabular}{|p{\textwidth}|}
\hline
\textbf{Example 1 }\\ \hline
\textbf{Target PET:} comfort women\\
\textbf{Sentence:} \textit{and what about the' comfort women industry in israel that uses slavic women as sex slaves} \\
\textbf{RankedPhrases:} {[}('sex slaves', 4.3283873945474625), ('slavic', 3.69672554731369), ('women', 1.4681523442268372), ('israel', 1.3728241324424744), ('comfort', 1.0837920159101486), ('industry', 0.8285560309886932){]} \\
\textbf{Failure:} The Target PET 'comfort women' was never identified as a MWE, and thus could not be detected. Additionally, polarized non-euphemisms like "sex slaves" are ranked higher as well as neutral candidates such as "slavic" or "women".  This is likely the result of highly polarized alternatives that produce a high score. \\ \hline
\end{tabular}
\end{table}
\begin{table}[h!]
\begin{tabular}{|p{\textwidth}|}
\hline
\textbf{Example 2} \\ \hline
\textbf{Target PET:} correctional facility \\
\textbf{Sentence:} \textit{very few correctional facilities have formal vocational education programs that provide offenders with marketable skills and assistance in employment planning} \\
\textbf{RankedPhrases:}{[}('offenders', 3.9866801872849464), ('vocational education programs', 2.453631855547428), ('very few', 2.0981270894408226), ('correctional facilities', 1.8522954508662224), ('marketable skills', 1.2003385424613953), ('assistance', 0.7983754873275757), ('employment', 0.5764055326581001), ('formal', 0.4696378782391548){]} \\
\textbf{Failure:} Here, again the Target PET was identified as a phase however the shift in sentiment was greater for the other phrases in the sentence and thus it was not ranked in one of the top two spots. \\ \hline
\end{tabular}
\end{table}
\begin{table}[h!]
\begin{tabular}{|p{\textwidth}|}
\hline
\textbf{Example 3} \\ \hline
\textbf{Target PET:} pro-life \\
\textbf{Sentence:} \textit{finally i think many pro-life people are politically naive and are too willing to accept empty promises }\\
\textbf{RankedPhrases:} {[}('politically naive', 8.384997591376305), ('empty promises', 4.581491872668266), ('pro-life', 4.001500993967056), ('people', 3.438477225601673), ('i think', 1.7039387673139572){]} \\
\textbf{Failure:} We count this example as a failure as our Target PET is in third place; however, we believe both of the top two candidates to be PETs. \\
\textbf{Interpretation: } politically naive: someone who has little knowledge and/or experience with politics and empty promises: promises made that are never intended to be carried out \\ \hline
\end{tabular}
\end{table}
\begin{table}[h!]
\begin{tabular}{|p{\textwidth}|}
\hline
\textbf{Example 4} \\ \hline
\textbf{Target PET:} expecting \\
\textbf{Sentence:} \textit{i had stopped searching while we were expecting our second child because we were unable to travel if called upon to candidate }\\
\textbf{RankedPhrases:} {[}('unable to travel', 7.015634283423424), ('searching', 1.7277799248695374), ('second child', 1.598520651459694), ('candidate', 0.5451297163963318){]} \\
\textbf{Failure:} The target PET is not a phrase candidate because it was incorrectly filtered out at the topic filtering stage. This is likely the case because "expecting" is an otherwise common word.  \\ \hline
\end{tabular}
\end{table}

\newpage
\section{New Applications}
\label{sec:newapps}

Below are a few examples where our algorithm shows promise for new applications.  We test our algorithm on sentences that are not in our corpus to see if it is able to detect PETs in unseen data. An example is shown below:

\begin{table}[h!]
\begin{tabular}{|p{\textwidth}|}
\hline
\textbf{Example 1 }\\ \hline
\textbf{Sentence:} \textit{i heard last week at her birthday party that she has a bun in the oven he whispered as he ate a hot dog bun} \\
\textbf{RankedPhrases:} {[}(\underline{'bun in the oven'}, 5.764157593250275), ('hot dog bun', 3.9777240827679634), ('he whispered', 3.6007840037345886), ('she has', 2.2385976165533066), ('party', 1.9190692454576492), ('he ate', 1.8731415495276451), ('her birthday', 1.3221752345561981){]} \\
\textbf{New PET:} bun in the oven \\
\textbf{Interpretation:} a baby in a belly; a pregnancy \\ \hline
\end{tabular}
\end{table}

Below, we show an example where our algorithm shows potential in distinguishing euphemistic versus non-euphemistic usages of the same word. First, we show the output for a non-euphemistic sentence containing a non-euphemistic usage of the PET \emph{dismissed}:

\begin{table}[h!]
\begin{tabular}{|p{\textwidth}|}
\hline
\textbf{Example 2 }\\ \hline
\textbf{Sentence:} \textit{the class is dismissed and we bow to each other expressing our gratitude for the shared experience} \\
\textbf{RankedPhrases:} {[}('shared experience', 3.9033331400714815), ('bow', 3.663858987390995), ('each other', 2.1924624936655164), (\underline{'dismissed'}, 1.9963299129158258), ('expressing', 1.848377185408026), ('class', 0.9816022356972098){]} \\
\textbf{Interpretation:} allowed to leave or disband \\ \hline
\end{tabular}
\end{table}

Now, we show the output for a sentence containing a euphemistic usage of \textit{dismissed}. Note how \emph{dismissed} is now detected as a euphemism, as well as its higher sentiment score compared to the previous example.
\begin{table}[h!]
\begin{tabular}{|p{\textwidth}|}
\hline
\textbf{Example 3 }\\ \hline
\textbf{Sentence:} \textit{at nichols college outside worcester massachusetts a non-tenured professor who questioned the leadership of the college president was summarily dismissed} \\
\textbf{RankedPhrases:} {[}(\underline{'dismissed'}, 5.921802910044789), ('was summarily', 3.158419349696487), ('worcester massachusetts', 1.4444764871150255), ('college', 1.196013430133462), ('non-tenured', 1.1229130360297859), ('president', 1.0259317518211901), ('leadership', 1.0157726714387536){]} \\
\textbf{Interpretation:} forced to leave a position; fired  \\ \hline
\end{tabular}
\end{table}

\end{document}